\documentclass{article}
\usepackage[preprint]{neurips_2021}
\usepackage[utf8]{inputenc} % allow utf-8 input
\usepackage[T1]{fontenc}    % use 8-bit T1 fonts
\usepackage{hyperref}       % hyperlinks
\usepackage{url}            % simple URL typesetting
\usepackage{booktabs}       % professional-quality tables
\usepackage{amsfonts}       % blackboard math symbols
\usepackage{nicefrac}       % compact symbols for 1/2, etc.
\usepackage{microtype}      % microtypography
\usepackage{xcolor}         % colors
\usepackage{multicol}
\usepackage{wrapfig}
\usepackage{subfig}
\usepackage{microtype}
\usepackage{subfig}
\usepackage{booktabs} % for professional tables
\usepackage{amsmath}
\usepackage{amsthm}
\usepackage{amssymb}
\usepackage{graphics}
\usepackage{mathtools,bm, stackengine}
\usepackage[algo2e, noend]{algorithm2e}
\newtheorem{proposition}{Proposition}

\newtheorem{lemma}{Lemma}
\newtheorem{remark}{Remark}
\usepackage{hyperref}
\usepackage{cleveref}
\usepackage{tabularx}
\usepackage{algorithm}% http://ctan.org/pkg/algorithms
\usepackage{algpseudocode}% http://ctan.org/pkg/algorithmicx
\usepackage[compact]{titlesec}
\titlespacing{\section}{1pt}{*0}{*0}
\titlespacing{\subsection}{1pt}{*0}{*0}
\titlespacing{\subsubsection}{1pt}{*0}{*0}
\usepackage{authblk}
\stackMath
\title{MUSBO: Model-based Uncertainty Regularized and \\  Sample Efficient  Batch Optimization for Deployment Constrained Reinforcement Learning }

\author[1]{DiJia Su}
\author[1]{Jason D. Lee}
\author[2]{John M. Mulvey}
\author[1]{H. Vincent Poor}
\affil[1]{Department of Electrical and Computer Engineering, Princeton University}
\affil[2]{Department of Operation Research and Financial Engineering, Princeton University}

\begin{document}
	
	\maketitle
	\begin{abstract}
		In many contemporary applications such as healthcare, finance, robotics, and recommendation systems, continuous deployment of new policies for data collection and online learning is either cost ineffective or impractical. We consider a setting that lies between pure offline reinforcement learning (RL) and pure online RL called deployment constrained RL in which the number of policy deployments for
		data sampling is limited. To solve this challenging task, we propose a new algorithmic learning framework called Model-based Uncertainty regularized and Sample Efficient Batch Optimization (MUSBO). Our framework discovers novel and high quality samples for each deployment to enable efficient data collection. During each offline training session, we bootstrap the policy update by quantifying the amount of uncertainty within our collected data. In the high support region (low uncertainty), we encourage our policy by taking an aggressive update. In the low support region (high uncertainty) when the policy bootstraps into the out-of-distribution region, we downweight it by our estimated uncertainty quantification. Experimental results show that MUSBO achieves state-of-the-art performance in the deployment constrained RL setting.
	\end{abstract}
	\section{Introduction}
	\label{submission}
	\vspace{-0.2cm}Recent advances in deep learning have enabled reinforcement learning (RL) to  achieve remarkable success in various applications \citep{silver2017mastering, openai2019solving,app1, app2}. However, despite RL's success, it suffers many problems. In particular, traditional RL algorithms require the agent to interact with the real world to collect large amounts of online data with the latest learned policy. However, the online deployment of the agent to the real world might be impossible in many real world applications \citep{matsushima2020deploymentefficient}. In the field of robotics or self-driving cars, for example, the cost of deploying the agent to the field may be too risky for the agent itself as well as its surrounding environment. In quantitative finance, trading strategy is usually carefully back-tested and calibrated offline with historical data and simulation. Since the market data has a low signal-to-noise ratio \citep{chen2020deep}, online training can easily lead the policy fits to the noise, which is dangerous and can potentially trigger unexpected large monetary loss.  In the recommendation system setting \citep{recommender}, after the policy has been trained offline, it will be deployed across different servers for serving many users. In such setting, online training of policy is difficult because the resulting policy might become unstable and/or cause bad user experience. Thus, large chunk of data is collected for the deployed policy. Then, the data is used for offline training. During the next scheduled deployment, the production level (data collection) policy is then updated. 
	
	\begin{figure}[!ht]
		\centering
		\includegraphics[width=0.7\textwidth]{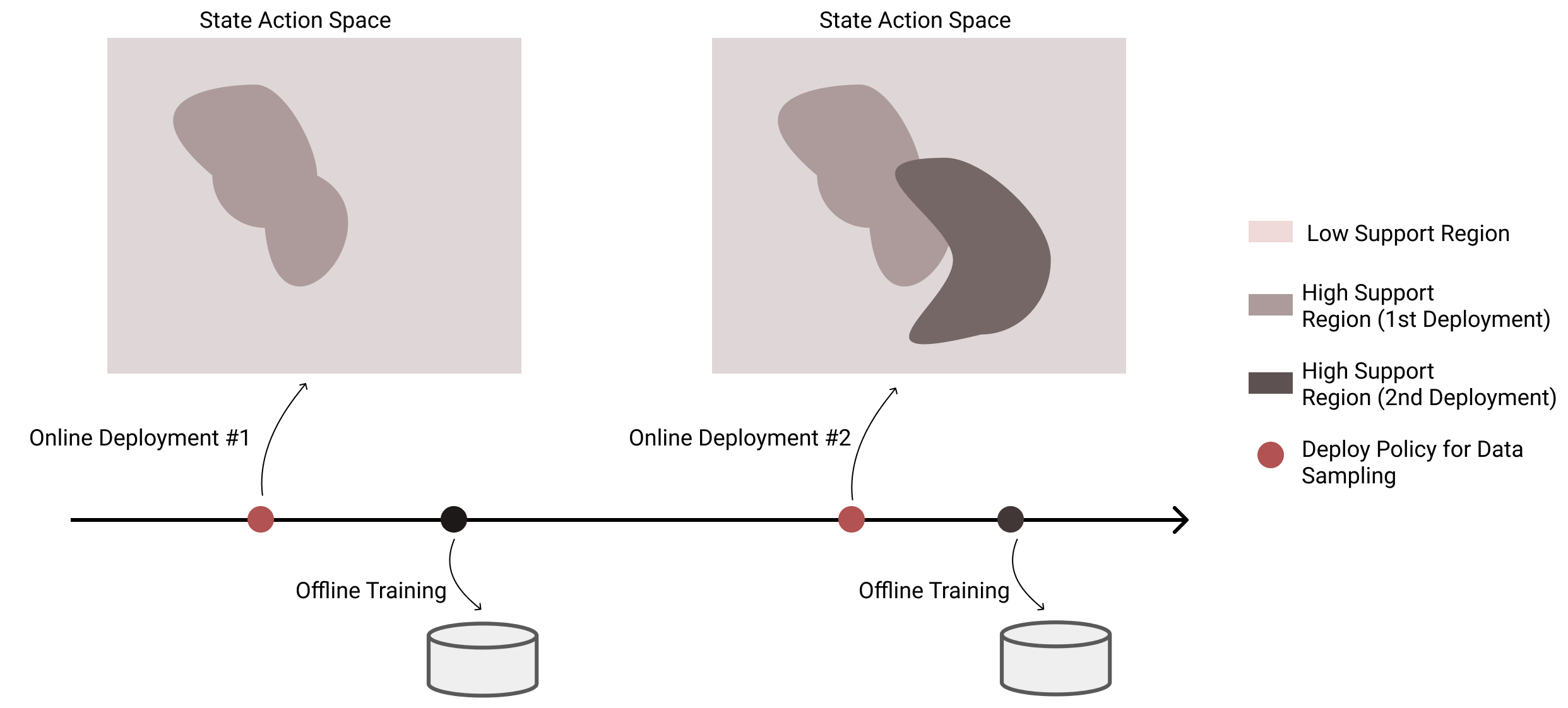}
		\caption{Illustration showing the deployment constrained RL setting. Large batch of data collection only occurs at the deployment point (showing two of them in red). The deployment and offline training occurs in an interleaved fashion. As the number of deployment increases, the state-action space will be explored more and more. Here, thicker-brown colored regions show the area of high support whereas the lighter color shows an under-explored (low support) region.}
		\label{fig:galaxy}
	\end{figure}
	To address this challenge, training an RL agent in an offline fashion seems to offer solution. In offline RL \citep{levine2020offline}, a static dataset is collected by a behavioral (or data collection) policy. Since the RL agent has no access to the online environment, training the RL agent faces many challenges. First, because the learning policy and the behavioral policy have different state visitation frequencies, evaluation of the offline policy is difficult. Second, during the training process, this discrepancy of the distribution might increase over the training time, a phenomenon known as distributional shift. Third, there might be a large extrapolation error when the value function is bootstrapped into out-of-distribution actions \citep{kumar2019stabilizing}, a phenomenon that can lead to learning divergence and instability. There is a large body of research work on pure offline RL. Despite their empirical success, since the pure offline RL environment is fundamentally different than our setting, we suspect developing novel deployment constrained RL can possibly achieve a large headroom of improvement, since offline RL algorithms assume zero interaction with the environment. 
	
	In this work, we present Model-based Uncertainty regularized and Sample Efficient Batch Optimization (MUSBO), a general framework for doing batch policy optimization in the  deployment constrained setting. To build an accurate model of the environment, our method encourages the data collection policy to \textit{discover high quality and novel data batches during each online deployment}. During the offline training, MUSBO samples fictitious rollouts from the learned model and performs policy update weighted with \emph{uncertainty regularized coefficient}, a term that quantifies the amount of uncertainty with respect to each state-action pair within the fictitious rollouts and the data. Our optimization process regularizes the update toward the high confidence regions. In areas of low confidence or low data support, MUSBO takes a pessimistic point of view and discounts the update by the uncertainty regularized coefficient when the policy bootstraps into out-of-distribution states and actions regions. We empirically compare our MUSBO to other strong baselines such as the state-of-the-art deployment constrained RL algorithms (BREMEN) \citep{matsushima2020deploymentefficient}, and we show that MUSBO is capable of achieving significant policy improvement while using smaller amounts of data and fewer deployment.  
	\section{Related Works}
	\vspace{-0.2cm}The related research literature can be broadly categorized into three categories: deployment-constrained RL, offline RL, and model-based RL. 
	
	\textbf{Deployment-constrained RL.} To the best of our knowledge, \cite{matsushima2020deploymentefficient} is the first work that proposed the concept of deployment-constrained efficiency. In their paper, the authors proposed the algorithm Behavior-Regularized Model-ENsemble (BREMEN) that enforces KL-divergence between the learning policy and the behavioral policy for learning update. They compared their approach across numerous baseline such as Soft-Actor-Critic (SAC)\citep{sac}, Model-based-TRPO (METRPO)\citep{kurutach2018modelensemble}, and show that their methodology provides the strongest empirical performance. Here in our paper, we compare our method MUSBO directly with a state-of-the-art method (BREMEN). Different from BREMEN, we first provide a theoretical analysis characterizing the improvement in terms of value function and optimization lower bounds. Second, our method is different because we emphasize at 1) propose to weight the contribution of each policy update  by  uncertainty quantification, 2) propose a method of measuring state action uncertainty by utilizing a new set of dynamics models with next state estimation error, 3) propose to make use of uncertainty quantification to maximize the discovery of novel data transitions during each deployment.
	
	On the other hand, there also some related research works such as \cite{bai2020provably} and in semi-batch RL such as \cite{semi1, semi2, semi3, semi4}.
	
	\textbf{Offline RL}. Unlike deployment-constrained setting, offline RL assumes no interaction with the environment. Thus, the learning policy needs to reason about the behavioral policy and make policy update base on that. The two most related literatures are \citet{mopo} and \citet{morel}. In \citet{mopo}, the authors proposed an uncertainty penalized MDPs in which the reward function was used to explicitly penalize the uncertainty. On the other hand, the authors from \citet{morel} proposed a pessimistic MDP that divides the environment into two regions: known or unknown. When the agent is entering into the unknown region, the MDP undergoes a halt state (or absorbing state) in which a large negative reward will be assigned to penalize this action. In both cases, the proposed uncertainty penalization comes from the reward function, which is a totally different setup than what is being proposed in our paper. Here, our method enables a regularization approach. Our method regularizes our policy update toward high confidence region while down-weights or regularized it away when the policy bootstraps into the unfamiliar  \emph{state and actions} regions. 
	
	Besides these two, there is also a large body of offline RL research. In model-free offline RL, the common techniques are either 1) enforcing the learning policy to stay close with the behavioral policy as in \citet{fujimoto2019offpolicy, kumar2019stabilizing, wu2019behavior, nachum2019algaedice, zhang2020gendice, nachum2019dualdice} or 2) ensembles of Q values for stabilizing the learning and behaviors as in  \citet{ghasemipour2021emaq, wu2019behavior, nair2020accelerating}.  
	
	\textbf{Model-Based RL}. In model-based approach, the world representation is learned first and then is used for generating imaginary rollouts. Related research works are \citet{chua2018deep, janner2019trust, luo2019algorithmic, JMLR:v9:munos08a}. However, direct application of MBRL methods into the offline setting can be challenging due to distribution shifts. Nevertheless, the closely related research is \citet{kurutach2018modelensemble}, in which an ensemble of estimated model dynamics is used for generating fictitious rollous for stabilizing effects.   Similar approaches have also been investigated by \citet{mb, mb1, mb2, mb3}.

	\subsection{Background}
	
	We consider a discounted, infinite horizon \emph{Markov decision process (MDP)}, and denote $\Omega = (S, A, M, r, \mu_0, \gamma)$, where $S$ is the state space, $A$ is the action space, $M(s'|s,a)$ is the state transition kernel of transiting to a next state $s'$ from state $s$ while taking action $a$, $r(s,a)$ is the reward function, $\gamma \in (0, 1)$ is the discounting factor, and $\mu_0$ is the initial state distribution.  
	
	In RL, the goal is to optimize a policy $\pi(a|s)$ such that the expected discounted return $J_M(\pi)$ is maximized: $J_M(\pi) = \underset{\pi, M, \mu_0}{\mathbf{E}} \sum_{t=0}^\infty \gamma^t r(s_t, a_t)$. The value function is defined as $V_M^{\pi} (s) = \underset{\pi, s_{t+1} \sim M(s_t, a_t)}{\mathbf{E}} \{ \sum_{t=0}^{\infty} \gamma^t r(s_t, a_t) | s_0 = s \}$. We denote the optimal policy $\pi^* = \text{argmax}_{\pi} J_M(\pi)$. We further define $\rho^{\pi}_M$ to be the discounted state visited by $\pi$ on $M$ such that $\rho^\pi_{M} = (1-\gamma) \sum_{t=0} \gamma^t p_{S^{\pi}_t}$, where $p_{S^{\pi}_t}$ is the distribution of state at time $t$.

	In the \textbf{deployment constrained} setting, there is a limitation on the number of times that the policy can be deployed online for data collection. In real world applications such as recommender system or robotic control, due to the associated cost with each deployment, it is desirable to minimize the number of deployments to as small as possible. Let $I$ be the number of times of deployment, and let $|B|$ be the size of a large batch of training data collected when a policy has been taken online (or deployed), then, the total size of data collected throughout the entire process is $I \times |B|$. Further, let $B^{(i)}$ be the batch of data collected during the $i_{th}$ deployment. In each batch $B$, we store the data transition $\{(s, a, r, s')\}$. Note the difference between the deployment constrained setting and the pure offline RL setting. In the pure offline RL setting, we consider it as a single batch of data with $I=1$, whereas in the deployment constrained setting, the agent will have the opportunity to interact with the environment by deploying its learning or other behavioral policy for data collection purposes. 
	
	In \textbf{Model-Based RL} (MBRL), the world dynamics are learned from the sampled data. We estimate the ground truth environment transition dynamics $M^*$ by learning a forward next state transition dynamics model  $\hat{M}: S \times A \rightarrow S$. Let $\hat{M}^{(i)}$ be the estimated dynamics bootstrapped from data batch $B^{(1)} \bigcup B^{(2)} \bigcup ... \bigcup B^{(i)}$, in which we assume the model is updated with the aggregated data batches up to the $i_{th}$ deployment. We denote by $\hat{M}^{(i)}_{\theta}$ the estimated dynamics learned by neural networks  parameterized by $\theta$.
	
	% \begin{algorithm}[H]
	% 	\setcounter{AlgoLine}{0}
	% 	\SetAlgoLined
	% 	\textbf{Given:} Size of data batches $B$, Number of deployments $I$\;
	% 	$D_{all} \leftarrow \{ \}$\\
	% 	\For{i = 1, 2, 3, ..., I $\text{of deployments}$}{
	% 		%instructions\;
	% 		$\text{deploy} \pi \text{ online for data collection}$\\ \nl
	% 		$D_{all} \leftarrow D_{all} \bigcup$ Collect-Data-Low-Support-Region($\pi$) \\ \nl
	% 		$\hat{M}_{\theta} \leftarrow \text{Learn approximate transition dynamics model }$ \\ \nl
	% 		%\hat{M}: S \times A \rightarrow S \text{ with } D_{all}\\ 
	% 		$\text{Train Uncertainty-Labeler} (D_{all})$ \\ \nl
	% 		$\pi \leftarrow \text{ Train with MBRL}(\text{Uncertainty-Labeler}, \hat{M}_{\theta}, D_{all})$ 
	
	% 	}
	% 	$\textbf{return} \pi$
	% 	\caption{MUSBO: Model-based Uncertainty Regularized and Sample Efficient Batch Optimization}
	% 	\label{alg:1}
	% \end{algorithm}
	\section{Algorithmic Framework}
	For the purpose of exposition, we start by presenting an idealized version of MUSBO algorithmic framework. Then, we describe a practical version that we have implemented and perform quite well. The detailed algorithm is summarized in Algorithm \ref{alg:1}. 
	
	The heart of the algorithm relies on uncertainty estimation. We postulate that by having more diversified samples (data coverage), we can learn a more accurate representation of the environment, which is the key ingredient of MBRL. To get the most out of each deployment, our algorithm tries to explore the under-represented region (the low support area of our collected data) so as to minimize the amount of uncertainty or maximize the information gain. After building an accurate model, we can use it to generate fictitious rollouts for policy training. The idea is that we want to weight the contribution of each policy update by taking the uncertainty quantification on the fictitious rollouts (each state action pair) by \textbf{uncertainty regularized coefficient} ($U_{\hat{M}^{(i)}, {M}^{*}}(s,a)$) which characterize the level of uncertainty results from model estimation error (\cref{eqn:u}).  For fictitious rollouts that we have low data support, we discount the policy update by $U_{\hat{M}^{(i)}, {M}^{*}}(s,a)$ and discourage the policy from bootstrapping into unknown regions. In doing this, we are \emph{regularizing} the policy update into high support/confidence regions. 
	
	\textbf{Theoretical Results}. All the proofs in this section are deferred to the appendix. In the deployment constrained setting, we are interested in iterative improvement after each deployment. 
	% We characterize iterative improvement by developing a lower bound for $V^{\pi} _M$ with $D(\cdot, \cdot)$ as the discrepancy between the $M^*$ and $\hat{M}$: 
	% \vspace{-0.2cm}\begin{equation}
	%     V^{\pi}_M \geq V^{\pi}_{\hat{M}} - D(\hat{M}, \pi)
	% \end{equation}
	% Let $u(\hat{M}, \pi)$ be a scalar such that:   $u(\hat{M}^{(i)}, \pi)  \triangleq  \frac{V^{\pi}_{\hat{M}} - D(\hat{M}^{(i)}, \pi)}{V^{\pi}_{\hat{M}}}$. 
	Let $\pi_{\text{ref}}$  be a reference policy that we will deploy to collect the data, $d(\cdot, \cdot)$ be the closeness of the two policies, $D(\cdot, \cdot)$ be the discrepancy between the $M^*$ and $\hat{M}$, we make the followings four assumptions:
	\noindent\begin{minipage}{0.5\textwidth}
		\begin{equation}
			\tag{A1}
			\small
			V^{\pi}_{M^*} \geq V^{\pi}_{\hat{M}} -  D_{\pi_{\text{ref}}, \delta} (\hat{M}, \pi),  \text{             s.t.}  d(\pi, \pi_{\text{ref}}) \leq \delta
			\label{A1}
		\end{equation} 
	\end{minipage}%
	\begin{minipage}{0.5\textwidth}\centering
		\begin{equation}
			\tag{A2}
			\small
			\hat{M} = M^* \implies D_{\pi_{\text{ref}}}( \hat{M}, \pi) = 0 
			\label{A2}
		\end{equation}
	\end{minipage}%
	\begin{equation}
	\tag{A3}
	\small
 \text{L-Lipschitz in } V^{\pi}_{\hat{M}} \text{ w.r.t  to some norm} \text{ such that} |V^{\pi}_{\hat{M}}(s) - V^{\pi}_{\hat{M}} (s')| \leq L \|s - s^{'}\|, \forall{s, s'}
	\label{A3}
\end{equation}\vspace{-0.2cm} \begin{equation}
		\tag{A4}
		\small
		D_{\pi_\text{ref}} (\hat{M}, \pi) \text{ is given by the form of } E_{\tau \sim \pi_\text{ref}, M^*} \{ f(\hat{M}, \pi, \tau) \}
		\label{A4}
	\end{equation}

	Let $\hat{M}^{(i)}$ be the model estimated from the data collected up to the $i_{th}$ deployment.	Adapted from \citet{schulman2017trust, luo2019algorithmic}:
	\begin{lemma} The difference of the value functions in-between each deployment is bounded by:
		\begin{equation}
			\small{
				|V^\pi_{\hat{M}^{(i+1)}}   - V^\pi_{\hat{M}^{(i)}} | \leq \kappa L  \underset{(s,a) \sim \rho^{\pi}_{\hat{M}^{(i+1)}}}{\mathbf{E}} (\|\hat{M} ^{(i+1)} (s, a) - \hat{M}^{(i)}(s, a) \|) 
			}
		\end{equation}
	\end{lemma}
	
We define the following function $U_{\hat{M}^{(i)}, {M}^{*}}(s,a)$: $S \times A \rightarrow \mathbb{R}$ as an \textbf{uncertainty regularized coefficient}:
\begin{equation}
	\small
	\begin{split}
		\underset{\substack{(s,a) \sim \rho^{\pi}_{\hat{M}^{(i)}}}}{\mathbf{E}}  \{ U_{\hat{M}^{(i)}, {M}^{*}}(s,a)\} \triangleq    g\Big(\kappa\underset{\substack{(s,a) \sim \rho^{\pi}_{\hat{M}^{(i)}}}}{\mathbf{E}}  \|{M} ^{*} (s, a) - \hat{M}^{(i)}(s, a) \|\Big)
	\end{split}
	\label{eqn:u}
\end{equation}
where $g(x)  \triangleq	(V^\pi_{\hat{M}^{(i)}} -  x) ({V^\pi_{\hat{M}^{(i)}}})^{-1} $, and  $\kappa\triangleq(1-\gamma)^{-1} \gamma$.

%\begin{equation}
%	\small
%	\begin{split}
%		\underset{\substack{(s,a) \sim \rho^{\pi}_{\hat{M}^{(i)}}}}{\mathbf{E}}  \{ U_{\hat{M}^{(i)}, \hat{M}^{(i)}}(s,a)\} \triangleq    \Big(V^\pi_{\hat{M}^{(i)}} - \kappa  \underset{\substack{(s,a) \sim \rho^{\pi}_{\hat{M}^{(i)}}}}{\mathbf{E}}  \{  \nu_{\hat{M}^{(i)}, \hat{M}^{(i)}}(s,a)  \}\Big) ({V^\pi_{\hat{M}^{(i)}}})^{-1}    
%	\end{split}
%	\label{eqn:u}
%\end{equation}
\begin{proposition}
	Let  $U_{\hat{M}^{(i)}, {M}^{*}}(s,a)$ be the uncertainty regularized coefficient defined by \cref{eqn:u}, then the performance of $\pi$ on the ground-truth $M^*$ is lower bounded by that of the estimated $\hat{M}^{(i)}$, weighted by $U_{\hat{M}^{(i)}, {M}^{*}}(s,a)$:
	\begin{equation}
		V^\pi_{{M}^{*}} \geq  V^\pi_{\hat{M}^{(i)}}\underset{\substack{(s,a) \sim \rho^{\pi}_{{M}^*}}}{\mathbf{E}} \{ U_{\hat{M}^{(i)}, {M}^{*}}(s,a)\} .
		\label{eqn:lowerbound}
	\end{equation}
	\label{cor:1}
\end{proposition}
\vspace{-0.5cm} Proposition.\ref{cor:1} says that in optimizing the lower bound (the RHS of \cref{eqn:lowerbound}  ) with $V^\pi_{\hat{M}^{(i)}}$, we can maximize the overall performance of $\pi$ on the real dynamics $M^*$.

\textbf{Interpretation of $U_{\hat{M}^{(i)},{M^*}} (s, a)$} (uncertainty regularized coefficient).  Here, the term can be interpreted as an uncertainty quantification measure as a result of \textit{model estimation error} between $\hat{M}^{(i)}$ and ${M^*}$, with $\hat{M}^{(i)}$ being learned from the collected (and limited amount of) data up to the $i_{th}$ deployment. On the regions where we have high data support, $\hat{M}^{(i)}(s,a)$ is close to ${M^*}(s,a)$, thus their discrepancy decreases and $U_{\hat{M}^{(i)},{M^*}} (s, a)$ approaches to unity. On the regions that are less certain, the discrepancy increases and thus $U_{\hat{M}^{(i)},{M^*}} (s, a)$ decreases.  
\begin{remark}
	If our estimated model $\hat{M}^{(i)}$ is reasonably good, so that the model estimation error $\kappa  \|{M} ^{*} (s, a) - \hat{M}^{(i)}(s, a) \|$ is smaller than $V^\pi_{\hat{M}^{(i)}}$, and that $V^\pi_{\hat{M}^{(i)}}$ is positive, then $U_{\hat{M}^{(i)},{M^*}} (s, a)$ is a scalar between 0 and 1, and is approximately proportional to the accuracy of the model estimation.
	\label{cor:2}
\end{remark}
%\emph{If our estimated model $\hat{M}^{(i+1)}$ is reasonably good, and that the model estimation error $ \|\hat{M} ^{*} (s, a) - \hat{M}^{(i+1)}(s, a) \|$ is smaller than $V^\pi_{\hat{M}^{(i+1)}}$, then $U_{\hat{M}^{(i+1)},{M^*}} (s, a)$ is a scalar between 0 and 1, and is approximately proportional to the accuracy of the model estimation}. 
	Thus to optimize $V^\pi_{M^*}$, we instead optimize its lower bound, which is the value function at the estimated model $V^\pi_{\hat{M}^{(i)}}$ weighted by the uncertainty regularized coefficient. We assign a higher weight at the regions of  high confidence, and a lower weight at the regions of low confidence.  Next, we utilize our result from \cref{eqn:lowerbound} for establishing our MUSBO lower bound optimization algorithm.
	% \begin{proof}
	%  See Lemma 4.3 of \cite{luo2019algorithmic} and we simply use \cref{eqn:u} for lower bound
	% \end{proof}
		\begin{algorithm}[H]
		\setcounter{AlgoLine}{0}
		\SetAlgoLined
		\textbf{Given:} Size of data batches $B$, Number of deployments $I$,
		$D_{all} \leftarrow \{ \}$\\
		\For{i = 1, 2, 3, ..., I $\text{of deployments}$}{
			%instructions\;
			$\text{deploy } \pi \text{ online for data collection}$\\ \nl \label{lst:line:1}
			$D_{all} \leftarrow D_{all} \bigcup$ Collect-Data-Low-Support-Region($\pi$) \\ \nl \label{lst:line:2}
			$\hat{M}_{\theta} \leftarrow \text{Learn approximate transition dynamics model }$ \\ \nl \label{lst:line:3}
			%\hat{M}: S \times A \rightarrow S \text{ with } D_{all}\\ 
			$\text{Train Uncertainty-Labeler} (D_{all})$ \\ \nl \label{lst:line:4}
			$\pi \leftarrow \text{ Train with MBRL}(\text{Uncertainty-Labeler}, \hat{M}_{\theta}, D_{all})$ as in \cref{section:mbrl}
			% 		$\phantom{Th.....} \text{                as in \cref{section:mbrl} }$ 
		}
		$\textbf{return} \pi$
		\caption{MUSBO: Model-based Uncertainty Regularized and Sample Efficient Batch Optimization}
		\label{alg:1}
	\end{algorithm}
	\subsection{Practical Implementation}
	In this section, we explain the practical implementation of \cref{alg:1} in detail. For readers convenience, we have labeled the line number in the algorithm, and we will refer to the line number as we explain below.
	% \subsection{Theoretical Analysis}.
	
	\textbf{Learning the transition dynamics} (\cref{lst:line:2}):
	For estimating the transition dynamics $\hat{M}_{\theta}$, we use an ensemble of $N$ deterministic dynamics models with multi-layer fully-connected perceptrons as in \citet{model1,model2} to reduce model bias. They are trained to predict the next state with $L_2$ loss: \vspace{-0.5cm}\begin{equation}
		\phantom{........}\underset{\theta}{\min} \frac{1}{|D_{all}|} \sum_{(s_t, a_t, s_{t+1}) \in D_{all}} \|s_{t+1} - \hat{M}_{\theta}(s_t, a_t)\|_2 ^ 2.
	\end{equation}
	Since we use the data collected up to $i_{th}$ deployment, we drop the $(i)$ superscript on $\hat{M}^{(i)}_{\theta}$ hereafter. 

	\vspace{-0.0cm}\textbf{Uncertainty-Labeler} (\cref{lst:line:3}). This module is responsible for characterizing the levels of uncertainty in the collected data and approximated $U_{\hat{M}^{(i)}, {M}^{*}}(s,a)$. Specifically, we want to identify the regions in our data that have high support or low support with respect to uncertainty.  Since the oracle is unavailable to us, here we can only approximate the uncertainty by state-actions visitation frequency. 
	% Unlike previous RL literatures which is allowed to train the model in the online fashion, in the deployment constrained setting, the model only has access to the previous batch of collected data ($B^{(1)}, .. B^{(i-1)}$). 
	Shall the state-actions visited frequently, more pairs of them will show up in the data. If we were to train models on the batches of data to predict the next state transition dynamics, then, our prediction will be more accurate on the states that we have seen (in the collected data), and less accurate on the unfamiliar state. Thus, we have established that the uncertainty quantification measure as the next state prediction error. 
	
	In the actual implementation, we have used $K$ ensembles of Probabilistic Neural Networks (PNN) \citep{chua2018deep} to \textit{capture the uncertainty} within the data. Let $\hat{P}_{\phi}: S \times A \rightarrow S $ be the PNN with parameter $\phi$. In PNN, the network has its output neurons parameterized by a Gaussian distribution in the effort of capturing the  uncertainty.  As explained above, we use $\hat{P}_{\phi}$ to quantify the uncertainty by prediction error. The lower the prediction error in relation to the ground truth, the higher the support. Our uncertainty-labeler approximates the uncertainty regularized coefficient $U_{\hat{M}^{(i)}, M^*} (s, a)$ by $\hat{U}(a,s)$ (define below shortly), and we have dropped the dependency on subscript since $\hat{U}$ is trained with the data collected up to the $i_{th}$ deployment.

	The actual implementation of $\hat{U}(a,s)$ is motivated by \cref{cor:2}. Let ${\hat{\tau}}$ be the fictitious trajectory generated by  $\hat{M}_{\theta}$ (the learned dynamics), we use $\hat{P}_{\phi}$ to quantify the uncertainty. Since the ground truth state is unavailable to us in the fictitious rollouts, to calculate the prediction error, we further approximate it by the intra-discrepancy error within the ensembles. We randomly sample two models (a, b) from $K$ Uncertainty-Labeler ensemble, for each state-action pairs in ${\hat{\tau}}$, we label them :
	\vspace{-0.1cm}\begin{eqnarray}
		\widehat{{U}}(s_t, a_t) = \text{exp}(-\alpha \|\hat{s}^{(a)}_{t+1} - \hat{s}^{(b)}_{t+1}\|_1),   \text{for } (s_t, a_t) \in \hat{\tau}
		\label{eqn:intra}
	\end{eqnarray}
	where $ \hat{s}^{(a)}_{t+1}$ = $\hat{P}_{\phi}^{(a)} ({s_t}, a_t)$ is the next state predicted by the sampled $a_{th}$ model from the Uncertainty-Labeler (and similarly for $b$ index), and $\alpha > 0$ is a temperature parameter.  Here, we have assumed that  our estimated models $\hat{P}_{\phi}$ are operating on the "reasonably good" regime (as by \cref{cor:2} ).

	Note that here, we have used different sets and different types of networks for estimating the model dynamics $\hat{M}_{\theta}$ and the uncertainty $\hat{P}_{\phi}$. We use $\hat{M}_{\theta}$ (deterministic networks) to generate fictitious rollouts, and we use $\hat{P}_{\phi}$ (PNN, a specialized uncertainty probabilistic network) as the "judge" to quantify the uncertainty for the generated fictitious rollouts. We separate out these two networks intentionally to avoid possible error propagation between the two modules. $\hat{P}_\phi$ is trained by (\cref{eqn:P}). For detail discussion, see Appendix F.
	
	\textbf{Collect Data from Low Support Region} (\cref{lst:line:1}). To maximize the benefit out of each deployment and to build an accurate representation $\hat{M}_\theta$ of the environment,  we emphasize that we want to achieve the maximal data coverage as well as exploring the under-explored region (the low support area). 
	
	Exploration in the deployment constrained setting is non-trivial because it is hard to evaluate which action-state pairs will fill the low support region and lead to high data coverage. During online data sampling, for each state that our agent encounters, we identify the amount of uncertainty by taking the actions which lead to maximal \emph{prediction error} between our predicted next state ($\hat{s}'$) and the ground truth next state (as by eqn.\ref{eqn:action} in the Appendix E).  Following a trajectory of maximal prediction error leads to novel experience discovery and reduce the number of unknown regions within the data.  In the ablation study (Fig.\ref{fig:ablation} a. and b.), we show that this strategy leads to more accurate learned dynamics $\hat{M}_\theta$. (For detailed implementation, please refer to Appendix E).
	
	% \begin{remark}
	% Let $f$ be a function whose derivative exists in every point, then $f$ 
	% is a continuous function.
	% \end{remark}

	\subsection{Offline Training with MBRL Method} \label{section:mbrl}
	Next, we define our offline model-based training method (\cref{lst:line:4}).  In the function below, we provide a practical instantiation of MUSBO offline model-based method for learning a policy.  
	% \begin{algorithm}
	%   \DontPrintSemicolon
	%   \SetKwFunction{FMain}{MUSBO MBRL Training}
	%   \SetKwProg{Fn}{Function}{:}{}
	%   \SetKwProg{Pn}{Function}{:}{\KwRet}
	%   \Pn{\FMain{$\hat{M}_{\theta}$, $D_{all}$, \text{Uncertainty-Labeler}}}{
	%     \nl Initialize $\pi_{\text{init}}$ with behavior cloning as given by eqn. \ref{eqn:bc_init} \\ \nl \label{lst:line:5}
	%     \For{\text{training iterations}}{
	%       \nl \label{lst:line:7}\text{Randomly sample dynamics} $\hat{M}_{\theta}$ from $N$ ensembles\\
	%         \For{\text{optimization steps}}{
	%             \nl \label{lst:line:8} $\hat{\tau} \leftarrow$ sample fictitious trajectory from $\hat{M}_{\theta}$\\ \nl\label{lst:line:9}
	%         $\hat{\tau}_{labeled} \leftarrow$  label fictitious trajectory with \phantom{..................}Uncertainty-Labeler ($\hat{\tau}$) as in eq.(\ref{eqn:intra})\\ \nl \label{lst:line:10}
	%             train with TRPO with $\hat{\tau}_{labeled}$, $\pi_{\text{init}}$ as in eq.(\ref{eqn:trpo}) 
	
	%         }
	%       }
	%   }
	%  \label{function:1}
	% \end{algorithm}
	Our method utilizes trust region policy optimization (TRPO) \citep{schulman2017trust} with fictitious trajectory generated with learned ensembles of transition dynamics $\hat{M}_{\theta}$ weighted by Uncertainty-Labeler (or uncertainty regularized coefficient).
	
	\begin{algorithm}
		\DontPrintSemicolon
		\SetKwFunction{FMain}{MUSBO MBRL Training}
		\SetKwProg{Fn}{Function}{:}{}
		\SetKwProg{Pn}{Function}{:}{\KwRet}
		\Pn{\FMain{$\hat{M}_{\theta}$, $D_{all}$, \text{Uncertainty-Labeler}}}{
			\nl Initialize $\pi_{\text{init}}$ with the data collection policy. \\ \nl \label{lst:line:5}
			\For{\text{training iterations}}{
				\nl \label{lst:line:7}\text{Randomly sample a dynamics model} from $N$ ensembles of $\hat{M}_{\theta}$\\
				\For{\text{optimization steps}}{
					\nl \label{lst:line:8} $\hat{\tau} \leftarrow$ sample fictitious trajectory from $\hat{M}_{\theta}$\\ \nl\label{lst:line:9}
					$\hat{\tau}_{labeled} \leftarrow$  label fictitious trajectory with Uncertainty-Labeler ($\hat{\tau}$) as in eq.(\ref{eqn:intra})\\ \nl \label{lst:line:10}
					train with uncertainty regularized TRPO with $\hat{\tau}_{labeled}$, $\pi_{\text{init}}$ as in eq.(\ref{eqn:trpo}) 
					
				}
			}
		}
		\label{function:1}
	\end{algorithm}
	
	\textbf{Fictitious trajectory generated from learned dynamics} (\cref{lst:line:7} to \cref{lst:line:8}). 
	After each deployment, the environment dynamics transitions are estimated by an ensemble of $\{\hat{M}_{\theta}\}_1 ^N$ trained using $D_{\text{all}}$. To generate a fictitious trajectory, we first randomly sample a model $j \in (1, 2, .. , N)$, and then we roll-out the trajectory ${\hat{\tau}}$ by running the learning policy $\pi$ with the next state as $\hat{s}_{t+1} = \hat{M}_{\theta_j} (\hat{s_t}, a_t=\pi(\hat{s_t}))$, for $t \in {1,..., T}$. 
	
	\textbf{TRPO training with Uncertainty} (\cref{lst:line:9} to \cref{lst:line:10}).
	Next, we train the policy with uncertainty regularized TRPO. In the offline setting, TRPO is trained using imaginary rollouts generated from the learned dynamics $\hat{M}_\theta$. Utilizing Prop \ref{cor:1}, we use TRPO to optimize the $V^{\pi}_{M^*}$ by improving its lower-bound (the RHS). Here, we replace $V^{\pi}_{\hat{M}^{(i)}}(s)$ by $A^{\pi_{\vartheta_k}}(s,a)$, and we approximated $U_{\hat{M}^{(i)}, M^*}(s,a)$ by $\widehat{{U}}(s,a)$ (eqn.\ref{eqn:intra}), with TRPO:{
		\begin{eqnarray}
			\underset{\vartheta}{\text{argmax }} E_{{\pi_\vartheta({s}), {s}, \hat{P}_{\phi}^{(a,b)}}, \hat{M}_{\theta}}  \{ \frac{\pi_\vartheta(a |s)}{\pi_{\vartheta_k}(a|s)} A^{\pi_{\vartheta_k}}(s,a) \widehat{{U}}(a,s) \} \nonumber\\ 
			\text{s.t.        }   
			E_{a \sim \pi_\vartheta({s}), {s}, \hat{P}_{\phi}^{(a,b)}, \hat{M}_{\theta}} \{ D_{KL}(
			\pi_\vartheta( \cdot | s) \| \pi_{\vartheta_k}( \cdot | s)
			)\} \leq \delta 
			\label{eqn:trpo}
		\end{eqnarray}
	}
	where we set $\pi_{\vartheta_0} = \pi_{\text{init}}$ as the initial policy of the TRPO at the first iteration (as a way to impose the constraint on $\pi_\text{ref}$). Here, $A^{\pi_{\vartheta_k}}(s,a)$ is the advantage function of policy $\pi_{\vartheta_k}$ following a fictitious trajectory generated by $\hat{M}_{\theta}$.  Here, when the policy bootstraps into out-of-distribution states and actions regions (low support area), we discount the update. In doing this,our optimization process regularizes the update toward the high confidence regions.
	
	% Note that \cref{eqn:trpo} is the second term of the RHS in \cref{eqn:lowerbound} except that we have replaced the value function with the advantage function. Here, we scale the advantage function with \emph{intra-Uncertainty estimate} because we want to \emph{discount} over-estimated advantage from low confidence (or unfamiliar) state-action pairs which will lead to learning a suboptimal policy \cite{ross2012agnostic}. 
	
	\section{Empirical Results}
	We empirically evaluate our proposed MUSBO algorithm with five continuous control benchmarks using the MuJOCO\footnote{http://www.mujoco.org/} physics simulator: Walker2d, Hopper, Half-Cheetah, Ant, and Cheetah-Run. 
	\begin{figure}[h]
		\centering
		\includegraphics[width=0.99\columnwidth]{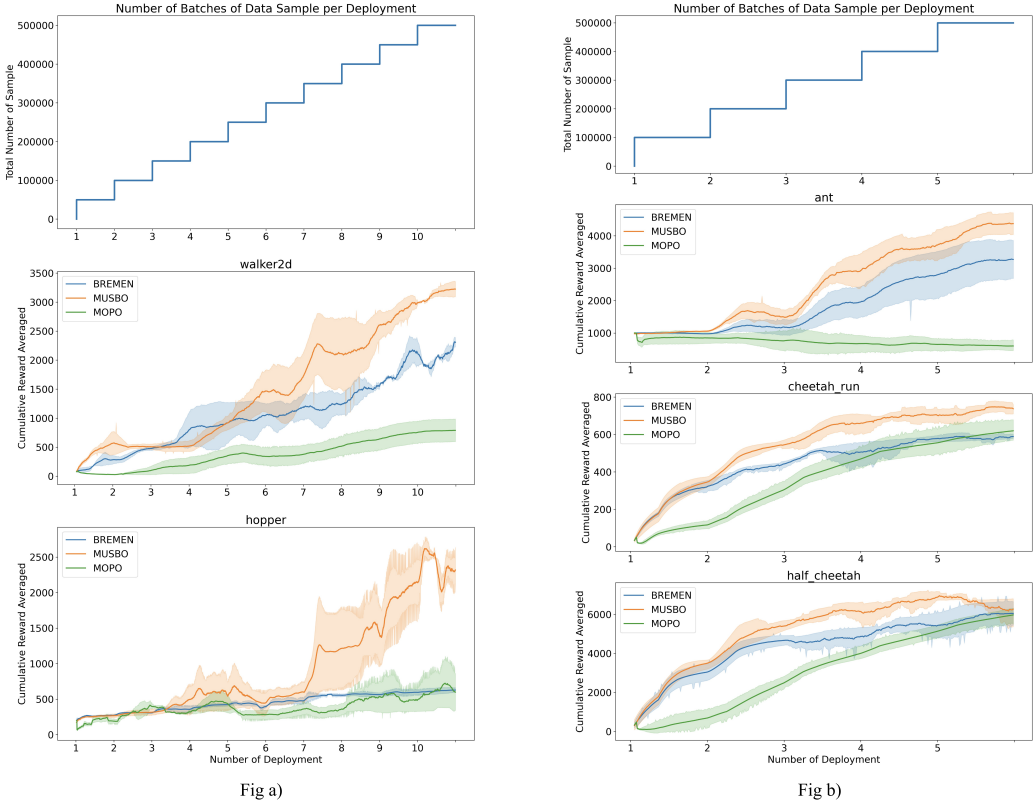}
		\caption{Empirical Evaluations of a) Walker2D and Hopper for $I=10$ deployments, $|B|=50k$, and b) Ant, Half\_Cheetah and Cheetah\_Run for $I=5$ deployments, $|B|=100k$. Total data consumption in both settings is $I \times |B| = 500k$. The x-axis is aligned showing the number of deployment.}
		\label{fig:500k1}
	\end{figure}
	\textbf{Baseline}. 
	We compare our MUSBO with BREMEN, a state-of-the-art algorithm\footnote{BREMEN has been compared against to SAC, Model-Ensembles-TRPO, BCQ, and BRAC (all adapted to deployment-constrained setting) and shows state-of-the-art performance.} that is designed specifically for deployment constrained setting. For BREMEN, we used the exact same hyper-parameters as what was originally proposed in \citet{matsushima2020deploymentefficient}. As for comparison, we also adapted MOPO \citep{mopo}, a pure-offline algorithm, to deployment-constrained setting. We used the latest learned policy for data collection. 
	
	\textbf{Evaluation Setup}. Following \citet{matsushima2020deploymentefficient}, our evaluation set up consisted of 5 deployments for Half-Cheetah, Ant, Cheetah-Run and 10 deployments for Walker2D and Hopper. To see the trade-off between the number of deployment versus data sample size, we perform our empirical tests on two separate cumulative data sizes of \emph{500k} and \emph{250k}. 
	
	For the settings of \emph{500k / (250k)} sample size experiments, the environments Half-Cheetah, Ant and Cheetah-Run have a per deployment data collection batch size $|B|=100k / (50k)$ (with 5 deployments in total). For the environments of  Walker2d and Hopper, they have a per deployment data collection batch size of $|B|=50k / (25k)$ (with 10 deployments in total). 
	
	% Similarly, for the settings of \emph{250k} sample size experiments, the environments Half-Cheetah, Ant and Cheetah-Run have a per deployment data collection batch size $B=50k$ (with 5 deployments in total). For the environments of  Walker2d and Hopper, they have a per deployment data collect batch size of $B=25k$ (with 10 deployments in total).
	
\begin{figure}[h]
		\centering
		\includegraphics[width=0.95\columnwidth]{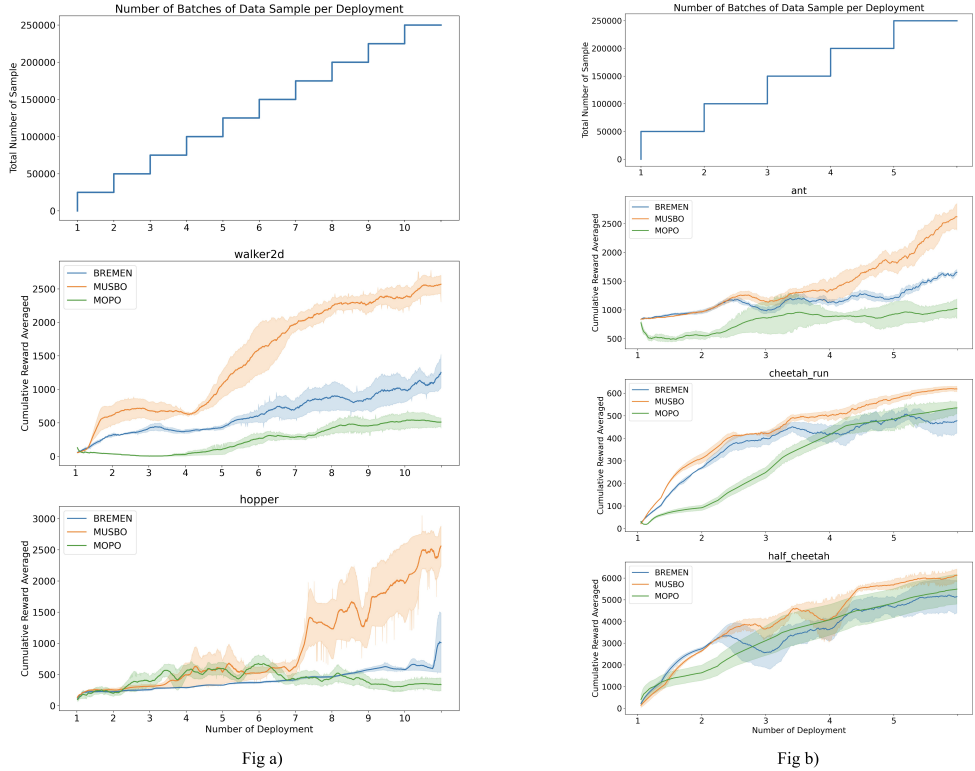}
		\caption{Empirical Evaluations of a) Walker2D and Hopper for $I=10$ deployments, $|B|=25k$ and b) Ant, Half\_Cheetah and Cheetah\_Run for $I=5$ deployments, $|B|=50k$. Total data consumption in both settings is $I \times |B| = 250k$. The x-axis is aligned showing the number of deployment.}
		\label{fig:250k}
	\end{figure}

	\textbf{Deployment Setup}. In the deployment constrained setting, the data collection only happens during the deployment. No training happens during this period. Only until a data batch size of $|B|$ has been collected, the agent will be taken offline for training and policy update.   At the first (initial) deployment, a random policy is used for data collection.   After that, the data collection will be replaced by the updated policy and will be launched to deploy again. We plot our 500k data size results in Fig.\ref{fig:500k1}, and we plot our 250k data size results on Fig.\ref{fig:250k}. For all experimental results, we averaged over 5 random seeds.
	% \begin{table}[htbp]
	%   \centering
	%   \resizebox{0.5\textwidth}{!}{\begin{minipage}{\textwidth}
	%         \begin{tabular}{lllll}
	% Deployment       & 2nd   & 3rd   & 4th   & 5th   \\
	% \toprule
	% MUSBO            & \textbf{0.994} & \textbf{0.963} & \textbf{0.945} & \textbf{0.944} \\
	% \midrule
	% Baseline(BREMEN) & 0.974 & 0.909 & 0.894 & 0.865 \\
	% \bottomrule
	%         \end{tabular}
	%                 \caption{ the amount of novelty of each data batches collected between each deployment. Novelty is measured as cosine distance between each observation (state) in $B^{(i)}$ versus the previous aggregated batches ($B^{(1)}..\bigcup B^{(i-1)}$), averaged over the number of transitions.}
	%         \label{Table:novely}
	%       \end{minipage}}
	% \end{table}
	\begin{table}[!htp]
		\centering
		\resizebox{0.5\columnwidth}{!}{%
			\begin{tabular}{lllll}
				Deployment       & 2nd   & 3rd   & 4th   & 5th   \\
				\toprule
				MUSBO            & \textbf{0.995} & \textbf{0.959} & \textbf{0.945} & \textbf{0.941} \\
				\midrule
				Baseline(BREMEN) & 0.972 & 0.903 & 0.891 & 0.858 \\
				\bottomrule
			\end{tabular}
		}
		\caption{Table showing the amount of novelty of each data batches collected between each deployment for cheetah\_run. Novelty is measured as cosine distance between each observation (state) in $B^{(i)}$ versus the previous aggregated batches ($B^{(1)}..\bigcup B^{(i-1)}$), averaged over the number of transitions.}
		\label{Table:novely}
	\end{table}
	% \begin{wrapfigure}{R}{0.45\columnwidth}
	%   \begin{center}
	%   \includegraphics[width=0.45\columnwidth]{./limage/abalation_horizontal.jpeg}
	%   \end{center}
	% \caption{Fictitious rollouts accuracy of our learned model dynamics for the Cheetah-Run environment, comparing between MUSBO and baseline (BREMEN). The vertical dotted lines show the point at which the deployments take place.
	% Comparing the fictitious rollouts generated from the learned model against the real rollouts from the envrionment, we show the energy distance on the left plot, and the trajectory-wise MSE on the right plot. Both plots (the \textbf{lower} the curve the better) show that our method MUSBO achieves learning a more accurate model. }
	%     \label{fig:model_accuracy}
	% \end{wrapfigure}
	
	\vspace{-0.5cm}\textbf{Empirical Details: 500k data size}. The top(first) figure of Fig.\ref{fig:500k1} shows the total sample size of each deployment on the y-axis, and on the x-axis, we show the number of deployments. We align along the x-axis for all figures. Our method works the best on the Walker2D and Hopper, in which our MUSBO approach achieves significant cumulative rewards especially in the longer deployments (after $6_{th}$). In the Hopper environment, we see that the baselines (BREMEN and MOPO) show incapable of learning a meaningful policy while our MUSBO significantly outperforms. On the other hand, in the Ant, Cheetah-Run and Half-Cheetah environment (as shown in part b. of Fig.\ref{fig:500k1}), our MUSBO is capable of achieving a higher performance using a smaller amount of data. For instance, in the Cheetah-Run environment, MUSBO already achieved 550 points in the second deployment but the baselines take four to five deployments to achieve the same score. 
	
	\textbf{Empirical Details: 250k data size}. Different than the previous set of experiments, here, we reduce the total data size by half to 250k. In general, when we reduce the data size, the performances of all algorithms will be reduced. Despite this, our MUSBO algorithm performs quite well even in this setting. Comparing to the 500k data size experiment, we also observe similar trends. In Fig.\ref{fig:250k} a), we plot the results for Walker2d and Hopper. For the former, we see a significant winning margin in Walker2d whereas in Hopper, the winning only happens in the $8_{th} \sim 9_{th}$ deployments. In Fig.\ref{fig:250k} b), we plot the results for Ant, Cheetah-Run and Half-Cheetah. We observe that our MUSBO algorithm achieves faster learning and stronger performance.
	
	\begin{figure}[!htp]
		\centering
		\includegraphics[width=0.99\columnwidth]{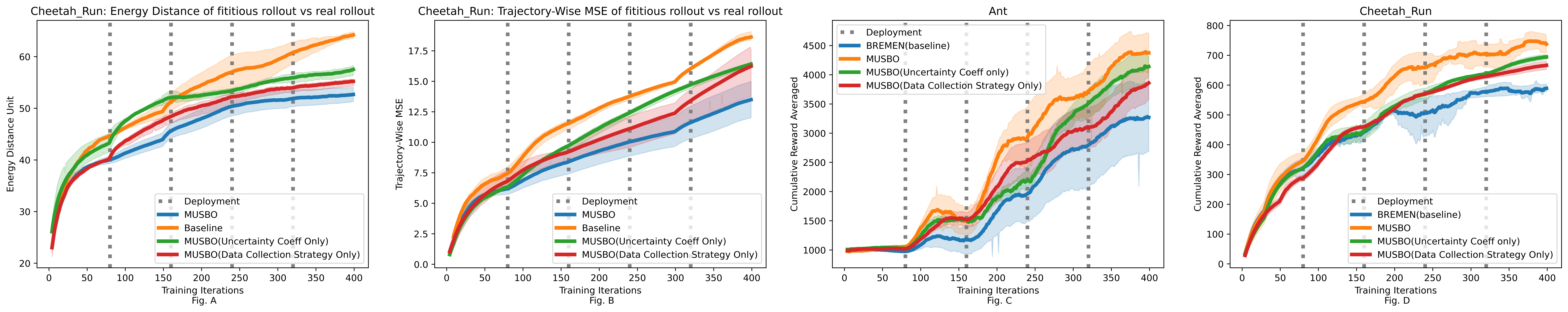}
		\caption{Ablation study on $I \times |B| = 500k$ setting: isolating the effects of having either uncertainty coeff only or  data collection strategy only (as in eqn.\ref{eqn:action}) on MUSBO.  On subplots a) and b) we show quality of the learned model dynamics. We compare the fictitious rollouts against real rollouts in terms of energy distance (the \textbf{lower} the better) and trajectory-wise MSE (the \textbf{lower} the better) for the cheetah\_run. We see that the data collection strategy gives stronger effect. We plot the cumulative rewards on fig. c) and d) for ant and cheetah\_run. We see that the effect of uncertainty coeff is stronger than the effect of data collection strategy. The combined effects of two components (which becomes MUSBO) gives the best performance not only on the learned dynamics (subplots a and b), also on the cumulative rewards (subplots c and d).}
		\label{fig:ablation}
	\end{figure}
	
	\section{Ablation Study}
	\label{sec:ablation}
	In this section, we examine the MUSBO algorithm in terms of the following three aspects: 1) discovery of high quality and novel data batches during each deployment; 2) whether this will lead to more accurate learning of model dynamics; 3) isolation effect of having uncertainty coefficient only or data-collection strategy only .
	To assess the novelty of data-batches during each deployment, we calculate the average cosine distance between the current batch ($B^{(i)}$) versus the aggregation of all of the previous batches ($B^{1} \bigcup B^{2} \bigcup ... B^{(i-1)}$) and then show the result in Table.\ref{Table:novely}. Note that deployment 1 is not shown because the initial data collection policy is a random policy. Our result shows that our MUSBO algorithm leads to much higher novel transitions discovery. 
	
	In Fig.\ref{fig:ablation} a) and b) subplots, we compare the performance of the fictitious rollouts from the learned model dynamics versus the rollouts from the real environment. We first plot the energy distance in a) and we show the mean square error (MSE) of trajectory-wise rollouts in b). As the number of deployment increases, our method is capable of discovering high quality transitions which result in a better estimated model.  
	
	Lastly, in Fig.\ref{fig:ablation}, we isolate each component and show the effects of including either 1) only the uncertainty coefficient or 2) only our data collection strategy as in eqn.\ref{eqn:action}. In terms of cumulative rewards, the former (coeff only) has a stronger effect than latter (data collection strategy only). In terms of the learned model dynamics, latter has a stronger effect.  Combining both gives the optimal effect on the cumulative rewards (subplots c and d) and leads learning an accurate model (subplots a and b).
	\section{Conclusion}
	\label{conclusion}
	% In this paper, we have proposed the algorithmic framework Model-based Uncertainty Regularized Sample efficient Batch Optimization (MUSBO) for optimizing the policy learning under the deployment-constrained setting. Through theoretical analysis, we first build a lower bound that utilizes uncertainty quantification of state-action pairs in the dataset. Our proposed strategy encourages novel data transition discovery during each deployment. Then, during offline training, we weight each state-action pair with uncertainty measures and discourage the policy update from bootstrapping into unknown (or low support) regions. Our empirical analysis shows that MUSBO outperforms the state-of-the-art algorithm in the MuJuco benchmarks.  
	%In this paper, we have proposed the algorithmic framework MUSBO for optimizing the policy learning under the deployment-constrained setting. It is possible to extend our framework (uncertainty regularized coefficient) to model-free algorithms, which hasn't been explored in this paper (and thus a limitation). Also, the data collection policy might subject to additional constraint (such as minimal performance requirement) which hasn't not been explored in our paper. One potential negative social impact is the carbon footprint from training MBRL methods. Training policy offline consumes energy for extensive amount of time might be harmful to the environment. 
		In this paper, we have proposed the algorithmic framework MUSBO for optimizing the policy learning under the deployment-constrained setting. One limitation is that we have assume  L-Lipschitz in the value function for our theoretical analysis which might not be the case in the practical setting. We have strong empirical result which justify it could be a viable solution.  One potential negative social impact is the carbon footprint from training MBRL methods. Training policy offline consumes energy for extensive amount of time might be harmful to the environment. One way to mitigate is to train a smaller network using fewer iteration but at the cost of performance.
	
	% \section*{Acknowledgements}
	
	% \textbf{Do not} include acknowledgements in the initial version of
	% the paper submitted for blind review.
	
	% If a paper is accepted, the final camera-ready version can (and
	% probably should) include acknowledgements. In this case, please
	% place such acknowledgements in an unnumbered section at the
	% end of the paper. Typically, this will include thanks to reviewers
	% who gave useful comments, to colleagues who contributed to the ideas,
	% and to funding agencies and corporate sponsors that provided financial

	\bibliographystyle{neurips_2021}
	\bibliography{main}

\begin{thebibliography}{37}
\expandafter\ifx\csname natexlab\endcsname\relax\def\natexlab#1{#1}\fi

\bibitem[{Bai et~al.(2020)Bai, Xie, Jiang, and Wang}]{bai2020provably}
Yu~Bai, Tengyang Xie, Nan Jiang, and Yu-Xiang Wang. 2020.
\newblock \href {http://arxiv.org/abs/1905.12849} {Provably efficient
  q-learning with low switching cost}.

\bibitem[{Chen et~al.(2020)Chen, Pelger, and Zhu}]{chen2020deep}
Luyang Chen, Markus Pelger, and Jason Zhu. 2020.
\newblock \href {http://arxiv.org/abs/1904.00745} {Deep learning in asset
  pricing}.

\bibitem[{Chu and Kitani(2020)}]{semi4}
Wen-Hsuan Chu and Kris~M. Kitani. 2020.
\newblock Neural batch sampling with reinforcement learning for semi-supervised
  anomaly detection.
\newblock In \emph{Computer Vision -- ECCV 2020}, pages 751--766, Cham.
  Springer International Publishing.

\bibitem[{Chua et~al.(2018)Chua, Calandra, McAllister, and
  Levine}]{chua2018deep}
Kurtland Chua, Roberto Calandra, Rowan McAllister, and Sergey Levine. 2018.
\newblock \href {http://arxiv.org/abs/1805.12114} {Deep reinforcement learning
  in a handful of trials using probabilistic dynamics models}.

\bibitem[{Ernst et~al.(2005)Ernst, Geurts, and Wehenkel}]{semi1}
Damien Ernst, Pierre Geurts, and Louis Wehenkel. 2005.
\newblock Tree-based batch mode reinforcement learning.
\newblock \emph{Journal of Machine Learning Research}, 6:503--556.

\bibitem[{Feinberg et~al.(2018)Feinberg, Wan, Stoica, Jordan, Gonzalez, and
  Levine}]{mb3}
Vladimir Feinberg, Alvin Wan, Ion Stoica, Michael~I. Jordan, Joseph~E.
  Gonzalez, and Sergey Levine. 2018.
\newblock \href {http://arxiv.org/abs/1803.00101} {Model-based value estimation
  for efficient model-free reinforcement learning}.

\bibitem[{Fujimoto et~al.(2019)Fujimoto, Meger, and
  Precup}]{fujimoto2019offpolicy}
Scott Fujimoto, David Meger, and Doina Precup. 2019.
\newblock \href {http://arxiv.org/abs/1812.02900} {Off-policy deep
  reinforcement learning without exploration}.

\bibitem[{Ghasemipour et~al.(2021)Ghasemipour, Schuurmans, and
  Gu}]{ghasemipour2021emaq}
Seyed Kamyar~Seyed Ghasemipour, Dale Schuurmans, and Shixiang~Shane Gu. 2021.
\newblock \href {http://arxiv.org/abs/2007.11091} {Emaq: Expected-max
  q-learning operator for simple yet effective offline and online rl}.

\bibitem[{Haarnoja et~al.(2018)Haarnoja, Zhou, Abbeel, and Levine}]{sac}
Tuomas Haarnoja, Aurick Zhou, Pieter Abbeel, and Sergey Levine. 2018.
\newblock \href {http://arxiv.org/abs/1801.01290} {Soft actor-critic:
  Off-policy maximum entropy deep reinforcement learning with a stochastic
  actor}.

\bibitem[{Jaakkola et~al.(1999)Jaakkola, Singh, and Jordan}]{semi3}
Tommi Jaakkola, Satinder Singh, and Michael Jordan. 1999.
\newblock Reinforcement learning algorithm for partially observable markov
  decision problems.
\newblock \emph{Advances in Neural Information Processing Systems}, 7.

\bibitem[{Janner et~al.(2019)Janner, Fu, Zhang, and Levine}]{janner2019trust}
Michael Janner, Justin Fu, Marvin Zhang, and Sergey Levine. 2019.
\newblock \href {http://arxiv.org/abs/1906.08253} {When to trust your model:
  Model-based policy optimization}.

\bibitem[{Kaiser et~al.(2020)Kaiser, Babaeizadeh, Milos, Osinski, Campbell,
  Czechowski, Erhan, Finn, Kozakowski, Levine, Mohiuddin, Sepassi, Tucker, and
  Michalewski}]{mb1}
Lukasz Kaiser, Mohammad Babaeizadeh, Piotr Milos, Blazej Osinski, Roy~H
  Campbell, Konrad Czechowski, Dumitru Erhan, Chelsea Finn, Piotr Kozakowski,
  Sergey Levine, Afroz Mohiuddin, Ryan Sepassi, George Tucker, and Henryk
  Michalewski. 2020.
\newblock \href {http://arxiv.org/abs/1903.00374} {Model-based reinforcement
  learning for atari}.

\bibitem[{Kidambi et~al.(2020)Kidambi, Rajeswaran, Netrapalli, and
  Joachims}]{morel}
Rahul Kidambi, Aravind Rajeswaran, Praneeth Netrapalli, and Thorsten Joachims.
  2020.
\newblock \href {http://arxiv.org/abs/2005.05951} {Morel : Model-based offline
  reinforcement learning}.

\bibitem[{Kumar et~al.(2019)Kumar, Fu, Tucker, and
  Levine}]{kumar2019stabilizing}
Aviral Kumar, Justin Fu, George Tucker, and Sergey Levine. 2019.
\newblock \href {http://arxiv.org/abs/1906.00949} {Stabilizing off-policy
  q-learning via bootstrapping error reduction}.

\bibitem[{Kurutach et~al.(2018{\natexlab{a}})Kurutach, Clavera, Duan, Tamar,
  and Abbeel}]{kurutach2018modelensemble}
Thanard Kurutach, Ignasi Clavera, Yan Duan, Aviv Tamar, and Pieter Abbeel.
  2018{\natexlab{a}}.
\newblock \href {http://arxiv.org/abs/1802.10592} {Model-ensemble trust-region
  policy optimization}.

\bibitem[{Kurutach et~al.(2018{\natexlab{b}})Kurutach, Clavera, Duan, Tamar,
  and Abbeel}]{model2}
Thanard Kurutach, Ignasi Clavera, Yan Duan, Aviv Tamar, and Pieter Abbeel.
  2018{\natexlab{b}}.
\newblock \href {http://arxiv.org/abs/1802.10592} {Model-ensemble trust-region
  policy optimization}.

\bibitem[{Lange et~al.(2012)Lange, Gabel, and Riedmiller}]{semi2}
Sascha Lange, Thomas Gabel, and Martin Riedmiller. 2012.
\newblock \href {https://doi.org/10.1007/978-3-642-27645-3_2} {Batch
  reinforcement learning}.
\newblock \emph{Reinforcement Learning: State of the Art}.

\bibitem[{Levine et~al.(2020)Levine, Kumar, Tucker, and Fu}]{levine2020offline}
Sergey Levine, Aviral Kumar, George Tucker, and Justin Fu. 2020.
\newblock \href {http://arxiv.org/abs/2005.01643} {Offline reinforcement
  learning: Tutorial, review, and perspectives on open problems}.

\bibitem[{Luo et~al.(2019)Luo, Xu, Li, Tian, Darrell, and
  Ma}]{luo2019algorithmic}
Yuping Luo, Huazhe Xu, Yuanzhi Li, Yuandong Tian, Trevor Darrell, and Tengyu
  Ma. 2019.
\newblock \href {http://arxiv.org/abs/1807.03858} {Algorithmic framework for
  model-based deep reinforcement learning with theoretical guarantees}.

\bibitem[{Matsushima et~al.(2020)Matsushima, Furuta, Matsuo, Nachum, and
  Gu}]{matsushima2020deploymentefficient}
Tatsuya Matsushima, Hiroki Furuta, Yutaka Matsuo, Ofir Nachum, and Shixiang Gu.
  2020.
\newblock \href {http://arxiv.org/abs/2006.03647} {Deployment-efficient
  reinforcement learning via model-based offline optimization}.

\bibitem[{Mishra et~al.(2017)Mishra, Abbeel, and Mordatch}]{model1}
Nikhil Mishra, Pieter Abbeel, and Igor Mordatch. 2017.
\newblock \href {http://arxiv.org/abs/1703.04070} {Prediction and control with
  temporal segment models}.

\bibitem[{Munos and Szepesv{{\'a}}ri(2008)}]{JMLR:v9:munos08a}
R{{\'e}}mi Munos and Csaba Szepesv{{\'a}}ri. 2008.
\newblock \href {http://jmlr.org/papers/v9/munos08a.html} {Finite-time bounds
  for fitted value iteration}.
\newblock \emph{Journal of Machine Learning Research}, 9(27):815--857.

\bibitem[{Nachum et~al.(2019{\natexlab{a}})Nachum, Chow, Dai, and
  Li}]{nachum2019dualdice}
Ofir Nachum, Yinlam Chow, Bo~Dai, and Lihong Li. 2019{\natexlab{a}}.
\newblock \href {http://arxiv.org/abs/1906.04733} {Dualdice: Behavior-agnostic
  estimation of discounted stationary distribution corrections}.

\bibitem[{Nachum et~al.(2019{\natexlab{b}})Nachum, Dai, Kostrikov, Chow, Li,
  and Schuurmans}]{nachum2019algaedice}
Ofir Nachum, Bo~Dai, Ilya Kostrikov, Yinlam Chow, Lihong Li, and Dale
  Schuurmans. 2019{\natexlab{b}}.
\newblock \href {http://arxiv.org/abs/1912.02074} {Algaedice: Policy gradient
  from arbitrary experience}.

\bibitem[{Nair et~al.(2020)Nair, Dalal, Gupta, and
  Levine}]{nair2020accelerating}
Ashvin Nair, Murtaza Dalal, Abhishek Gupta, and Sergey Levine. 2020.
\newblock \href {http://arxiv.org/abs/2006.09359} {Accelerating online
  reinforcement learning with offline datasets}.

\bibitem[{OpenAI et~al.(2019{\natexlab{a}})OpenAI, :, Berner, Brockman, Chan,
  Cheung, Debiak, Dennison, Farhi, Fischer, Hashme, Hesse, Jozefowicz, Gray,
  Olsson, Pachocki, Petrov, de~Oliveira~Pinto, Raiman, Salimans, Schlatter,
  Schneider, Sidor, Sutskever, Tang, Wolski, and Zhang}]{app2}
OpenAI, :, Christopher Berner, Greg Brockman, Brooke Chan, Vicki Cheung,
  Przemyslaw Debiak, Christy Dennison, David Farhi, Quirin Fischer, Shariq
  Hashme, Chris Hesse, Rafal Jozefowicz, Scott Gray, Catherine Olsson, Jakub
  Pachocki, Michael Petrov, Henrique~Ponde de~Oliveira~Pinto, Jonathan Raiman,
  Tim Salimans, Jeremy Schlatter, Jonas Schneider, Szymon Sidor, Ilya
  Sutskever, Jie Tang, Filip Wolski, and Susan Zhang. 2019{\natexlab{a}}.
\newblock \href {http://arxiv.org/abs/1912.06680} {Dota 2 with large scale deep
  reinforcement learning}.

\bibitem[{OpenAI et~al.(2019{\natexlab{b}})OpenAI, Akkaya, Andrychowicz,
  Chociej, Litwin, McGrew, Petron, Paino, Plappert, Powell, Ribas, Schneider,
  Tezak, Tworek, Welinder, Weng, Yuan, Zaremba, and Zhang}]{openai2019solving}
OpenAI, Ilge Akkaya, Marcin Andrychowicz, Maciek Chociej, Mateusz Litwin, Bob
  McGrew, Arthur Petron, Alex Paino, Matthias Plappert, Glenn Powell, Raphael
  Ribas, Jonas Schneider, Nikolas Tezak, Jerry Tworek, Peter Welinder, Lilian
  Weng, Qiming Yuan, Wojciech Zaremba, and Lei Zhang. 2019{\natexlab{b}}.
\newblock \href {http://arxiv.org/abs/1910.07113} {Solving rubik's cube with a
  robot hand}.

\bibitem[{Peska and Vojtas(2020)}]{recommender}
Ladislav Peska and Peter Vojtas. 2020.
\newblock \href {https://doi.org/10.1145/3372923.3404781} {Off-line vs. on-line
  evaluation of recommender systems in small e-commerce}.
\newblock In \emph{Proceedings of the 31st ACM Conference on Hypertext and
  Social Media}, HT '20, page 291–300, New York, NY, USA. Association for
  Computing Machinery.

\bibitem[{Schulman et~al.(2017)Schulman, Levine, Moritz, Jordan, and
  Abbeel}]{schulman2017trust}
John Schulman, Sergey Levine, Philipp Moritz, Michael~I. Jordan, and Pieter
  Abbeel. 2017.
\newblock \href {http://arxiv.org/abs/1502.05477} {Trust region policy
  optimization}.

\bibitem[{Silver et~al.(2017)Silver, Hubert, Schrittwieser, Antonoglou, Lai,
  Guez, Lanctot, Sifre, Kumaran, Graepel, Lillicrap, Simonyan, and
  Hassabis}]{silver2017mastering}
David Silver, Thomas Hubert, Julian Schrittwieser, Ioannis Antonoglou, Matthew
  Lai, Arthur Guez, Marc Lanctot, Laurent Sifre, Dharshan Kumaran, Thore
  Graepel, Timothy Lillicrap, Karen Simonyan, and Demis Hassabis. 2017.
\newblock \href {http://arxiv.org/abs/1712.01815} {Mastering chess and shogi by
  self-play with a general reinforcement learning algorithm}.

\bibitem[{Veerapaneni et~al.(2020)Veerapaneni, Co-Reyes, Chang, Janner, Finn,
  Wu, Tenenbaum, and Levine}]{mb2}
Rishi Veerapaneni, John~D. Co-Reyes, Michael Chang, Michael Janner, Chelsea
  Finn, Jiajun Wu, Joshua~B. Tenenbaum, and Sergey Levine. 2020.
\newblock \href {http://arxiv.org/abs/1910.12827} {Entity abstraction in visual
  model-based reinforcement learning}.

\bibitem[{Vinyals et~al.(2019)Vinyals, Babuschkin, Czarnecki, Mathieu, Dudzik,
  Chung, Choi, Powell, Ewalds, Georgiev, Oh, Horgan, Kroiss, Danihelka, Huang,
  Sifre, Cai, Agapiou, Jaderberg, Vezhnevets, Leblond, Pohlen, Dalibard,
  Budden, Sulsky, Molloy, Paine, Gulcehre, Wang, Pfaff, Wu, Ring, Yogatama,
  W{\"u}nsch, McKinney, Smith, Schaul, Lillicrap, Kavukcuoglu, Hassabis, Apps,
  and Silver}]{app1}
Oriol Vinyals, I.~Babuschkin, W.~Czarnecki, Micha{\"e}l Mathieu, Andrew Dudzik,
  J.~Chung, D.~Choi, R.~Powell, Timo Ewalds, P.~Georgiev, Junhyuk Oh, Dan
  Horgan, Manuel Kroiss, Ivo Danihelka, Aja Huang, L.~Sifre, Trevor Cai,
  John~P. Agapiou, Max Jaderberg, A.~S. Vezhnevets, R{\'e}mi Leblond, Tobias
  Pohlen, Valentin Dalibard, D.~Budden, Yury Sulsky, James Molloy, T.~L. Paine,
  Caglar Gulcehre, Ziyu Wang, T.~Pfaff, Yuhuai Wu, Roman Ring, Dani Yogatama,
  Dario W{\"u}nsch, Katrina McKinney, O.~Smith, T.~Schaul, T.~Lillicrap,
  K.~Kavukcuoglu, Demis Hassabis, Chris Apps, and D.~Silver. 2019.
\newblock Grandmaster level in starcraft ii using multi-agent reinforcement
  learning.
\newblock \emph{Nature}, pages 1--5.

\bibitem[{Wang et~al.(2019)Wang, Bao, Clavera, Hoang, Wen, Langlois, Zhang,
  Zhang, Abbeel, and Ba}]{wang2019benchmarking}
Tingwu Wang, Xuchan Bao, Ignasi Clavera, Jerrick Hoang, Yeming Wen, Eric
  Langlois, Shunshi Zhang, Guodong Zhang, Pieter Abbeel, and Jimmy Ba. 2019.
\newblock \href {http://arxiv.org/abs/1907.02057} {Benchmarking model-based
  reinforcement learning}.

\bibitem[{Wu et~al.(2019)Wu, Tucker, and Nachum}]{wu2019behavior}
Yifan Wu, George Tucker, and Ofir Nachum. 2019.
\newblock \href {http://arxiv.org/abs/1911.11361} {Behavior regularized offline
  reinforcement learning}.

\bibitem[{Yu et~al.(2020)Yu, Thomas, Yu, Ermon, Zou, Levine, Finn, and
  Ma}]{mopo}
Tianhe Yu, Garrett Thomas, Lantao Yu, Stefano Ermon, James Zou, Sergey Levine,
  Chelsea Finn, and Tengyu Ma. 2020.
\newblock \href {http://arxiv.org/abs/2005.13239} {Mopo: Model-based offline
  policy optimization}.

\bibitem[{Zhang et~al.(2019)Zhang, Vikram, Smith, Abbeel, Johnson, and
  Levine}]{mb}
Marvin Zhang, Sharad Vikram, Laura Smith, Pieter Abbeel, Matthew~J. Johnson,
  and Sergey Levine. 2019.
\newblock \href {http://arxiv.org/abs/1808.09105} {Solar: Deep structured
  representations for model-based reinforcement learning}.

\bibitem[{Zhang et~al.(2020)Zhang, Dai, Li, and Schuurmans}]{zhang2020gendice}
Ruiyi Zhang, Bo~Dai, Lihong Li, and Dale Schuurmans. 2020.
\newblock \href {http://arxiv.org/abs/2002.09072} {Gendice: Generalized offline
  estimation of stationary values}.

\end{thebibliography}

	%%%%%%%%%%%%%%%%%%%%%%%%%%%%%%%%%%%%%%%%%%%%%%%%%%%%%%%%%%%%

	%%%%%%%%%%%%%%%%%%%%%%%%%%%%%%%%%%%%%%%%%%%%%%%%%%%%%%%%%%%%

	\appendix
	% \section{Do \emph{not} have an appendix here}
	
	% \textbf{\emph{Do not put content after the references.}}
	% %
	% Put anything that you might normally include after the references in a separate
	% supplementary file.
	
	% We recommend that you build supplementary material in a separate document.
	% If you must create one PDF and cut it up, please be careful to use a tool that
	% doesn't alter the margins, and that doesn't aggressively rewrite the PDF file.
	% pdftk usually works fine. 
	
	% \textbf{Please do not use Apple's preview to cut off supplementary material.} In
	% previous years it has altered margins, and created headaches at the camera-ready
	% stage. 
	\newpage
	\newpage
	\clearpage
	\section{Proof of Lemma 1}
	Adapted from \cite{schulman2017trust, luo2019algorithmic}, we let $W_j$ be the expected return when executing $\hat{M}^{(i+1)}$ for the first $j$ steps, and then switch to $\hat{M}^{(i)}$ for the remaining steps. 
	\begin{proof}
		\begin{equation}
			W_j = \underset{\substack{a_t \sim \pi(s_t) \\ \forall{j >t \geq 0, s_{t+1} \sim \hat{M}^{(i+1)} (\cdot | s_t, a_t)}  \forall{t \geq j, s_{t+1} \sim \hat{M}^{(i)}} (\cdot | s_t, a_t)}}{\mathbf{E}} \{ \sum_{t=0} \gamma^t r(s_t, a_t) | s_0 = s \}
		\end{equation}
		
		Thus, we have $W_0 = V^{\pi}_{\hat{M}^{(i)}}$, and $W_{\infty} =  V^{\pi}_{\hat{M}^{(i+1)}}$. Next, we write:
		
		\begin{equation}
			V^{\pi} _{\hat{M}^{(i+1)}} - V^{\pi} _{\hat{M}^{(i)}} = \sum_{j=0}^\infty (W_{j+1} - W_{j})
			\label{eqn:proof1}
		\end{equation}
		
		We expand $W_j$ and $W_{j+1}$ so that we can cancel the shared terms:
		\begin{equation*}
			\begin{split}
				W_j = R_j +\underset{s_{j}, a_j \sim \pi, \hat{M}^{(i+1)}}{{\mathbf{E}}} \{     
				\underset{s_{j+1}\sim \hat{M}^{(i)}(\cdot | s_t, a_t)}{{\mathbf{E}}} \{ \gamma^{j+1} V^{\pi}_{\hat{M}^{(i)}}(s_{j+1})\}\}  
			\end{split}
		\end{equation*}
		\begin{equation*}
			\begin{split}
				W_{j+1} = R_j + \underset{s_{j}, a_j \sim \pi, \hat{M}^{(i+1)}}{{\mathbf{E}}} \{     
				\underset{s_{j+1}\sim \hat{M}^{(i+1)}(\cdot | s_t, a_t)}{{\mathbf{E}}} \{ \gamma^{j+1} V^{\pi}_{\hat{M}^{(i)}}(s_{j+1})\}\}  
			\end{split}
		\end{equation*}
		% \mathclap{\substack{
		% \begin{equation}
		% \begin{split}
		%      W_{j+1} &= R_j + \\&\stackunder{s_{j}, a_j \sim \pi, \hat{M}^{(i+1)}}{{\mathbf{E}}} \{     
		%         \underset{s_{j+1}\sim \pi, \hat{M}^{(i+1)}(\cdot | s_t, a_t)}{{\mathbf{E}}} \{ \gamma^{j+1} V^{\pi}_{\hat{M}^{(i+1)}}(s_{j+1})\}\}  
		% \end{split}
		% \end{equation}
		
		% \begin{equation}
		% \begin{split}
		%      W_{j+1} &= R_j + \\&\mathclap{\substack{s_{j}, a_j \sim \pi, \hat{M}^{(i+1)}}{{\mathbf{E}}} \{     
		%         \stackunder[5pt]{s_{j+1}\sim \pi, \hat{M}^{(i+1)}(\cdot | s_t, a_t)}{{\mathbf{E}}} \{ \gamma^{j+1} V^{\pi}_{\hat{M}^{(i+1)}}(s_{j+1})\}\}}  
		% \end{split}
		% \end{equation}
		where $R_j$ is the expected return of the first $j$ time step. Next, we cancel the share terms so that: 
		\begin{equation}
			\small
			\begin{split}
				W_{j+1} - W_j = \gamma^{j+1} \underset{s_j, a_j \sim \pi, \hat{M}^{(i+1)}}{\mathbf{E}}\bigl\{  \underset{\substack{s' \sim \hat{M}^{(i+1)}(\cdot | s_j, a_j)} }{\mathbf{E}}  \{ V^{\pi}_{\hat{M}^{(i)}} (s')\}- 
				\underset{\substack{s' \sim \hat{M}^{(i)}(\cdot | s_j, a_j)} }{\mathbf{E}}\{V^{\pi}_{\hat{M}^{(i)}} (s') \}  \bigr\}
			\end{split}
			\label{eqn:proposition2}
		\end{equation}
		
		%  \\ \hat{S}_{j+1} \sim \hat{M}^{(i)}(\cdot | S_j, A_j)
		% \begin{align*}
		% \renewcommand\useanchorwidth{T}
		%     &W_{j+1} - W_j = \gamma^{j+1} \stackunder[5pt]{\mathbf{E}}{s_j, a_j \sim \pi, \hat{M}^{(i+1)}}\bigl\{  \\& \underset{\substack{\hat{S}'_{j+1} \sim \hat{M}^{(i+1)}(\cdot | S_j, A_j) \\ \hat{S}_{j+1} \sim \hat{M}^{(i)}(\cdot | S_j, A_j)} }{\mathbf{E}}  \{ V^{\pi}_{\hat{M}^{(i)}} (\hat{S}'_{j+1}) - V^{\pi}_{\hat{M}^{(i)}} (\hat{S_{j+1}}) \}  \bigr\}
		% \end{align*}
		
		% \begin{align*}
		% \renewcommand\useanchorwidth{T}
		%   \hat{\boldsymbol{\eta}} = \stackunder[5pt]{\text{arg max }}{
		%   ojbk}
		%   x_1 -j
		% \end{align*}

		Thus, based on \cref{eqn:proposition2}, we have:
		% \begin{equation}
		% \begin{split}
		% 			V^\pi_{\hat{M}^{(i+1)}} -  V^\pi_{\hat{M}^{(i)}}  = 
		% 			(1-\gamma)^{-1} \gamma \underset{\substack{ A \sim \pi \\ S \sim {\rho^{\pi}_{\hat{M}^{(i+1)}}}}}{\mathbf{E}} \bigl\{ \\ \underset{\hat{s}' \sim \hat{M}^{(i)}(\cdot | s, a)}{\mathbf{E}} \{V^{\pi}_{\hat{M}^{(i)}} (\hat{s'})\} - \underset{\hat{s}' \sim {\hat{M}^{(i+1)}}(\cdot | s, a)}{\mathbf{E}} \{V^{\pi}_{\hat{M}^{(i)}} (\hat{s'})\}  \bigr\}
		% \end{split}
		% \end{equation}
		\begin{equation}
			\begin{split}
				V^\pi_{\hat{M}^{(i+1)}} -  V^\pi_{\hat{M}^{(i)}}  = 
				\kappa \underset{s_j, a_j \sim \pi, \hat{M}^{(i+1)}}{\mathbf{E}}\bigl\{  \underset{\substack{s' \sim \hat{M}^{(i+1)}(\cdot | s_j, a_j)} }{\mathbf{E}}  \{ V^{\pi}_{\hat{M}^{(i)}} (s')\}- 
				\underset{\substack{s' \sim \hat{M}^{(i)}(\cdot | s_j, a_j)} }{\mathbf{E}}\{V^{\pi}_{\hat{M}^{(i)}} (s') \}  \bigr\}
			\end{split}
			\label{eqn:tmp}
		\end{equation}
%		Multiply \cref{eqn:tmp} by $-1$ on both side and substitute in \cref{eqn:d}, thus we have \begin{equation}
%			\begin{split}
%				V^\pi_{\hat{M}^{(i+1)}}  \geq  V^\pi_{\hat{M}^{(i)}} - \qquad \kappa  \underset{(s,a) \sim \rho^{\pi}_{\hat{M}^{(i+1)}}}{\mathbf{E}}  \{  \nu_{\hat{M}^{(i)}, \hat{M}^{(i+1)}}(s,a)  \}.
%			\end{split}
%		\end{equation}
%		Since we have assumed the Lipschitzeness of $V^{\pi}_{\hat{M}}$, we can rewrite the LHS of \cref{eqn:tmp} that: \begin{equation}
%	\begin{split}
%			|V^\pi_{\hat{M}^{(i+1)}} -  V^\pi_{\hat{M}^{(i)}} |\leq  L|\hat{M}^{(i+1)}(s,a) - \hat{M}^{(i)}(s,a) |
%	\label{eqn:b_triangle}
%	\end{split}
%\end{equation}
 
 Let $\nu(s,a) =  \underset{\substack{s' \sim \hat{M}^{(i+1)}(\cdot | s_j, a_j)} }{\mathbf{E}}  \{ V^{\pi}_{\hat{M}^{(i)}} (s')\}- 
 \underset{\substack{s' \sim \hat{M}^{(i)}(\cdot | s_j, a_j)} }{\mathbf{E}}\{V^{\pi}_{\hat{M}^{(i)}} (s') \}$, 
 since we have assumed the Lipschitzeness of $V^{\pi}_{\hat{M}}$, we can bound $|\nu(s,a)| \leq L|\hat{M}^{(i+1)}(s,a) - \hat{M}^{(i)}(s,a)|$, then combine with triangle inequality, we have: 		\begin{equation}
 	\small{
 		|V^\pi_{\hat{M}^{(i+1)}}   - V^\pi_{\hat{M}^{(i)}} | \leq \kappa L  \underset{(s,a) \sim \rho^{\pi}_{\hat{M}^{(i+1)}}}{\mathbf{E}} (\|\hat{M} ^{(i+1)} (s, a) - \hat{M}^{(i)}(s, a) \|) 
 	}
 \label{eqn:lemma1_result}
 \end{equation}
		
	\end{proof}

\section{Proof of Proposition 1}
	\begin{proof}
Starting from eqn.\ref{eqn:tmp}, we substitute with $V^\pi_{M^*}$ and  $V^\pi_{\hat{M}^{(i)}}$,  and multiple both side with -1.  Due to the Lipschitzeness assumption, we can then bound $|\nu(s,a)|$ (with the corresponding substitution) on eqn.\ref{eqn:tmp} and by the definition of $U_{\hat{M}^{(i)}, {M}^{*}}(s,a)$, thus we have:
		\begin{equation}
	V^\pi_{{M}^{*}} \geq  V^\pi_{\hat{M}^{(i)}}\underset{\substack{(s,a) \sim \rho^{\pi}_{{M}^*}}}{\mathbf{E}} \{ U_{\hat{M}^{(i)}, {M}^{*}}(s,a)\} .
	\label{eqn:lowerbound_}
\end{equation}
\end{proof}

\clearpage
\addtocounter{section}{2}
	\section{Exploration to Collect Data from Low Support Region} 
	During each deployment, we want to collect the data from the low support (or un-visited) regions.  we make use of the uncertainty labeler to guide exploration to the un-visited regions for novel data discovery. This is achieved by injecting the ${\zeta}(a, s)$ as an exploration noise with the zero-mean normal distribution: $\mathcal{N}(0, \sigma=\zeta(a, s))$, where $\zeta$ is:
	\begin{equation}
		\zeta(a, s) = \underset{i \in \{\hat{P}_{\phi}\}_i^K}{\max} (\|s_{t+1} - \hat{s}_{t+1}\|_1)
	\end{equation}
	where $\hat{s}_{t+1}$ is the predicted next state and we take the maximum prediction error of the model within $ \{\hat{P}_{\phi}\}_i^K$ ensemble.
	In the Ablation Study (Section.\ref{sec:ablation}), we show that this exploration strategy leads to novel data discovery, and also contribute to better learned model dynamics. 
	
	Specifically, the action will be parameterized by a stochastic Gaussian policy (with parameter $\mu_\vartheta$) as:
	\begin{equation} \tag{16}
		a_t = \text{tanh} (\mu_\vartheta (s_t)) + \epsilon_{\text{const}} + \epsilon_{\zeta}
		\label{eqn:action}
	\end{equation}
	where $\epsilon_{\text{const}}$ and $\epsilon_{\zeta} $ are:
	% $\epsilon_{\text{fix}} \sim \mathcal{N}(\mu=0, \sigma=0.01)$ and $\epsilon_{\hat{U}} \sim \mathcal{N}(\mu=0, \sigma =\hat{U}(a=\text{tanh} (\mu_\vartheta (s_t)) + \epsilon_{\text{fix}}, s = s_t))$.
	\begin{equation*} 
		\begin{split}
			& \epsilon_{\text{const}} \sim \mathcal{N}(\mu=0, \sigma=0.01) \\
			& \epsilon_{\zeta} \sim \mathcal{N} \Bigl(\mu=0, \sigma =\zeta\bigl(a=\text{tanh} (\mu_\vartheta (s_t)) + \epsilon_{\text{const}}, s = s_t\bigr)\Bigr)
		\end{split}
	\end{equation*}
	
	Here, $\epsilon_{\text{const}}$ is an additive noise with a constant variance of 0.01, and on top of this, we also added another Gaussian noise with variance equal to $\zeta(a,s)$ from the uncertainty labeler to guide exploration.
	
	% We also tried $\hat{\hat{U}}(a,s)$ instead of $\hat{U}(a,s)$  for exploration noise, and we observed similar performance
	
	\section{Detailed Implementation}
	We used ADAM as the optimizer with a learning rate of 1e-3 for the model dynamics $\hat{T}$ with ensembles size of $N=5$. For the uncertainty-labeler, we use ensembles of PNN with $K=3$ and a learning rate of 1e-3. For the behavioral cloning, we used a learning rate of 5e-4. For all the collected data, we divided them into 85\% for training, and 15\% for validation (for model validation). The $\hat{T}, \hat{P}$ are trained with early stopping when their performance on the validation set no longer improves after consecutive 3 episodes. We used this same setting for the behavioral cloning as well. 
	
	\textbf{Model Architecture}
	For our policy network, we parameterized it by two layers of fully-connected neural network with hidden units of 200. For the $\hat{T}$ model dynamics, we parameterized it by two layers of fully-connected neural network with hidden units of 1024.  We used this same configuration with $\hat{P}$ uncertainty-labeler and implemented with PNN. 
	
	\textbf{Training Time}
	The overall training time differs for each environment. We train all models on Nvidia T4 GPU. For the $500k$ data size experiment, the entire training duration (1 run) for Walker2D and Hopper environments are 18 hours and 26 hours respectively. For the Half-Cheetah, Ant, and Cheetah-Run environment, it is a a lot more faster. It takes 7 hours, 12 hours, and 8 hours per run, respectively.  For the $250k$ data size experiments, the training time is about 25 minutes faster than the $500k$ experiments.

	% \textbf{Hyper-parameters}
	Since we are applying Dyna-style update with neural network based dynamics models, following \citet{wang2019benchmarking} \citet{matsushima2020deploymentefficient}, we used the following reward functions for our dynamics models (for model-based training only) as:
	\begin{itemize}
		% \centering
		\item Walker2d, Hopper: $\dot{x_t} - 0.001\lVert a_t\rVert^2_2 + 1$
		\item CheetahRun: $\max(0, \min(\dot{x_t} /10, 1)) $
		\item Ant: $\dot{x_t} - 0.1\lVert a_t\rVert^2_2 - 3.0(z_t -0.57)^2 +1$
		\item HalfCheetah: $\dot{x_t} - 0.1\lVert a_t\rVert^2_2 $
	\end{itemize}
	
	We enabled termination in rollouts for the Hopper and Walker2D environments, and disabled that of the Ant, HalfCheetah, and CheetahRun environments (with a maximum step of 1000 for each episode). For the CheetahRun task, we adopt it from the DM control suit\footnote{\text{https://github.com/deepmind/dm\_control}}.
	
	\textbf{Other Hyper-parameters}
	We searched $\alpha$ over the set of \{0.28, 0.028, 0.0028\} and we used the same $\alpha = 0.028$ (the temperature parameter for eq.(\ref{eqn:intra})) for all environments. Similarly, for the action parameterization \cref{eqn:action}, we used the same constant $\sigma=0.01$ (variance term of $\epsilon_{\text{const}}$) for all environments.  The $\sigma$ of $\epsilon_{\text{const}}$ is searched over the set of \{0.01, 0.05, 0.1\}.

	We searched the rollout length on $\{250, 1000\}$, and the $\delta$ on $\{0.01, 0.05, 0.1\}$.  We summarized these three parameters ($L$, Rollouts Length, $\delta$) as above in \cref{tabl:hyper}.
	
	For discount factor $\gamma$ and GAE $\lambda$, we used the same set of hyperparmaters as in \citet{wang2019benchmarking}. Specifically, we used the same $\gamma = 0.99$ for all environment. Also, we used GAE $\lambda = 0.95$ for all environment except for Ant which has a GAE $\lambda = 0.97$.

	\begin{table}[]
		\centering
		\begin{tabular}{l|l|l|l}
			\toprule
			& $L$   & Rollouts Length & TRPO's $\delta$ \\
			\midrule
			Ant         & 2,000 & 250             & 0.05                                                       \\
			HalfCheetah & 2,000 & 250             & 0.1                                                        \\
			Hopper      & 6,000 & 1,000           & 0.05                                                       \\
			Walker2d    & 2,000 & 1,000           & 0.05                                                       \\
			CheetahRun  & 2,000 & 250             & 0.05    \\
			\bottomrule 
			
		\end{tabular}
		\caption{Hyper-parameters for MUSBO Algorithm}
		\label{tabl:hyper}
	\end{table}
	
	\textbf{Hyper-parameters for Baseline}. 
	For the BREMEN, we used the exact parameters as \cite{matsushima2020deploymentefficient}. For MOPO, we adapted it to the deployment setting. We used the latest learned policy for deployment, and then launched it to collect data batch of size $|B|$. For the CheetahRun task, we used the same set of parameters of HalfCheetah. On all environments, we tried hyper-parameters search on the ranges as originally proposed by the paper \citet{mopo}, we didn't find any improvement over the same set of parameters as originally proposed. Thus, we used the same set of parameters as originally proposed.

	\textbf{Training of the Probabilistic Neural Networks (PNN)}. 
	In PNN \citep{chua2018deep}, the network has its output neurons parameterized by a Gaussian distribution in the effort of capturing the uncertainty. This module is trained to minimize the following loss:\begin{equation} \tag{17}
		\begin{split}
			\text{loss}_{PNN} (\phi) = \sum_{n=1} ^{K} \{\mu_{\phi} (s_n, a_n) - s_{n+1}\}^{\top} \Sigma_{\phi}^{-1} (s_n, a_n) (\mu_{\phi} (s_n, a_n) - s_{n+1}) + \text{log det} \Sigma_{\phi}(s_n, a_n)
		\end{split}
	\label{eqn:P}
	\end{equation} where $\phi$ is the neural network learning parameters, $\mu_{\phi}$ and  $\Sigma_{\phi}$ are the mean and variance of the Gaussian distribution.
	
%	\textbf{Stabilizing Training}. 
%	To stabilize the training, in the actual implementation of $\widehat{U}(a_t, s_t)$ which approximates $U_{\hat{M}^{(i)}, M^*} (s, a)$, we dropped the division of the value function. Appendix D of our paper shows that the difference in value functions (numerator of eqn.\ref{eqn:u}) can be bounded by one-step model prediction error (under assumptions). Based on this, we further simplify the implementation of uncertainty labeler down to an one-step prediction error and incorporate it with \emph{functions} which mimics the behavior of the actual $U_{\hat{M}^{(i)}, M^*} (s, a)$. Note that when $\hat{M}_\theta$ and $M^*$ becomes close to  each other, $U_{\hat{M}^{(i)}, M^*} (s, a)$ approaches to unity. In our implementation, we have shown this similar behavior by incorporating exponential for $\widehat{U}$ (eqn.\ref{eqn:intra}). This \emph{approximation} gets rid of the division of value function which caused training \emph{instability}. We found this implementation works better empirically. 
%	
%	
	
	%%%%%%%%%%%%%%%%%%%%%%%%%%%%%%%%%%%%%%%%%%%%%%%%%%%%%%%%%%%%%%%%%%%%%%%%%%%%%%%
	%%%%%%%%%%%%%%%%%%%%%%%%%%%%%%%%%%%%%%%%%%%%%%%%%%%%%%%%%%%%%%%%%%%%%%%%%%%%%%%

\end{document}